%% file: main.tex
\documentclass{article} 
\usepackage{iclr2027_conference,times}
\iclrfinalcopy
\input{math_commands.tex}

\usepackage{graphicx}
\usepackage{hyperref}
\usepackage{booktabs} 
\usepackage{multirow} 
\usepackage{url}

\title{Rethinking Reasoning Paths as Phase-Structured Trajectories}

\author{%
  {\normalfont \textbf{Zhenghao He}\textsuperscript{$\dagger$}\quad
  \textbf{Guangzhi Xiong}\quad
  \textbf{Sanchit Sinha}}\\
  \textbf{Bohan Liu}\quad
  \textbf{Wenqian Ye}\quad
  \textbf{Aidong Zhang}\textsuperscript{$\dagger$}\\[5pt]
  Department of Computer Science, University of Virginia\\
  \textsuperscript{$\dagger$}\texttt{\{zhenghao, aidong\}@virginia.edu}
}

\usepackage{hyperref}       
\usepackage{url}            
\usepackage{booktabs}       
\usepackage{amsfonts}       
\usepackage{nicefrac}       
\usepackage{microtype}      
\usepackage{xcolor}         
\usepackage[table]{xcolor}
\usepackage{amsmath}
\usepackage{makecell}
\usepackage{graphicx}
\usepackage{multirow}
\usepackage{wrapfig}
\usepackage{subcaption}
\definecolor{myblue}{RGB}{220,230,242}  
\definecolor{mygreen}{RGB}{226,240,217} 
\DeclareCaptionLabelFormat{figsub}{Figure~\thefigure#2}

\begin{document}

\maketitle

\input{sec/abstract}

\input{sec/intro}

\input{sec/related_work}
\input{sec/method}
\input{sec/exp}
\input{sec/conclusion}

\bibliography{iclr2027_conference}
\bibliographystyle{iclr2027_conference}

\input{sec/appendix}


\end{document}

%% file: math_commands.tex
\usepackage{amsmath,amsfonts,bm}

\def\eqref#1{equation~\ref{#1}}

\def\1{\bm{1}}

\DeclareMathAlphabet{\mathsfit}{\encodingdefault}{\sfdefault}{m}{sl}
\SetMathAlphabet{\mathsfit}{bold}{\encodingdefault}{\sfdefault}{bx}{n}



%% file: sec/abstract.tex
\begin{abstract}
Large language models often improve problem-solving performance by generating multi-step reasoning paths, yet how to analyze the hidden states along these paths remains unclear. Existing approaches typically assign each intermediate state the final-answer correctness label and train probes across heterogeneous questions. We argue that this protocol obscures reasoning dynamics in two ways: (1) correctness prediction can exploit question-level variation rather than path quality, and (2) states aligned by absolute step indices may correspond to different functional phases of reasoning. 
In this work, we propose to view reasoning paths as phase-structured trajectories within fixed questions. We instantiate this view as \textbf{PAIR}, short for \textbf{P}hase-\textbf{A}ligned \textbf{I}ntra-question \textbf{R}easoning. PAIR samples multiple trajectories for each question, maps variable-length paths into shared relative phases based on normalized trajectory progress, and compares successful and unsuccessful trajectories only within the same question and phase. This yields phase-specific path-quality directions that better isolate path-quality signals from question-level variation.
Empirically, we find that standard across-question correctness probes lose much of their predictive power under within-question evaluation, suggesting that these probes partly rely on question-level information.
PAIR improves within-question trajectory ranking and Best-of-\(N\) trajectory selection across models and benchmarks. 
Phase-wise steering further shows that the learned directions can change generation outcomes, providing causal evidence that they capture trajectory-relevant information. 
\end{abstract}

%% file: sec/intro.tex
\section{Introduction}
\label{sec:intro}
Large language models can improve their problem-solving performance by generating multi-step reasoning paths before producing final answers. 
This behavior is most explicit in chain-of-thought prompting~\citep{wei2022chain}, where models are encouraged to write intermediate reasoning steps, and has been shown to improve performance on mathematical reasoning~\citep{cobbe2021gsm8k}, symbolic reasoning, and code generation tasks~\citep{chen2021evaluating}. 
The accuracy gains from this style of generation are substantial, but the internal mechanism by which it succeeds or fails remains opaque. 
As a model generates a reasoning path, \emph{what information about the eventual correctness is encoded in its hidden states, and how does this information evolve along the trajectory? }

\begin{figure}
    \centering
    \includegraphics[width=\linewidth]{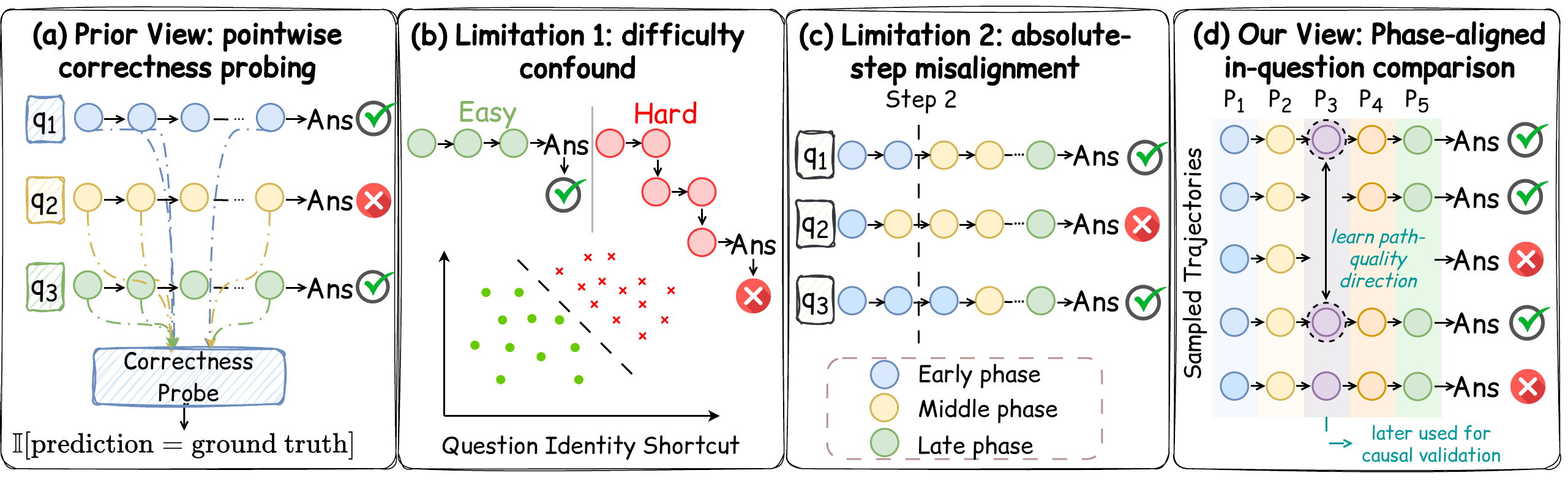}
    \caption{
Overview of our motivation.
(a) Prior probing labels intermediate states by final-answer correctness across different questions.
(b) Across-question prediction can exploit question identity rather than path quality.
(c) Absolute step indices can misalign reasoning phases across questions.
(d) We instead align trajectories by phase and compare paths within the same question.
}
\vspace{-10pt}
    \label{fig:intro}
\end{figure}

Recent work has increasingly analyzed language model reasoning through internal representations, using hidden-state probes, representation geometry, and activation directions to uncover task-relevant or correctness-related signals \citep{burns2212discovering,li2023inference,he2026reasoningchainofthoughtlatentcomputational}. 
In many such analyses, intermediate hidden states are treated as pointwise representations labeled by the final outcome, e.g., $y=\mathbb{I}[\hat{a}=a^\ast]$, and a probe or scorer is trained to predict final correctness from each state (Fig~\ref{fig:intro}a).

However, such across-question correctness prediction can conflate path quality with question-level variation (Fig~\ref{fig:intro}b). 
When examples are pooled across heterogeneous questions, the correctness label depends not only on the generated trajectory, but also on properties of the question being solved. 
This makes early correctness signals hard to interpret: strong prediction before substantial reasoning has occurred may reflect question-level information rather than path-level progress.

A second issue concerns how intermediate states are compared along the trajectory. 
Final correctness is a trajectory-level outcome, but pointwise probing assigns this same label to every intermediate state along the path. Recent work has begun to study reasoning as a structured trajectory, showing that step-specific activations can form separable subspaces and that correct and incorrect trajectories diverge at later stages \citep{sun2026llm}. 
These results suggest that reasoning has meaningful temporal geometry. 
However, existing trajectory analyses often align states by explicit step markers or absolute step indices. 
This creates a structural mismatch: ``Step 2'' can mean setup in one solution and the main calculation in another (Fig~\ref{fig:intro}c). 

In this work, we view reasoning paths as phase-structured trajectories within fixed questions (Fig.~\ref{fig:intro}d). 
A useful comparison should hold the question fixed, so that question-level information cannot serve as the main shortcut, and should compare states that occupy comparable positions within their trajectories. 
Under this view, the relevant question is not whether a hidden state predicts final correctness across a heterogeneous dataset, but whether one trajectory state is better than another for the same question and at the same relative phase. 

We instantiate this view as PAIR, short for \textbf{P}hase-\textbf{A}ligned \textbf{I}ntra-question \textbf{R}easoning. 
PAIR samples multiple reasoning trajectories for each question, maps variable-length paths to a shared set of phase slots, and constructs phase-matched comparisons between successful and unsuccessful trajectories from the same question. 
From these comparisons, PAIR learns phase-specific directions that rank successful states above unsuccessful states. 
We then use these directions for analysis and steering, testing whether they can actively redirect generation.

Empirically, we find that standard across-question correctness signals weaken substantially under within-question evaluation, suggesting that part of their apparent strength comes from question-level information. 
We further show that phase alignment reveals trajectory-quality structure that is obscured by absolute-step comparisons. 
The learned phase-specific directions improve within-question trajectory ranking and Best-of-\(N\) trajectory selection. 
Finally, phase-wise steering shows that these directions can change generation outcomes, providing causal evidence that they capture trajectory-relevant information.

Our contributions are as follows:
\begin{itemize}
    \item We diagnose key ambiguities in pointwise correctness analysis, showing that across-question probes can capture question-level variation and that absolute step indices can misalign reasoning stages.

    \item We introduce PAIR, a phase-aligned intra-question framework that compares successful and unsuccessful trajectories from the same question at matched relative phases to learn path-quality directions.

    \item We demonstrate that PAIR improves trajectory ranking and Best-of-\(N\) selection, and use phase-wise steering to provide causal evidence that the learned directions can affect reasoning generation.
\end{itemize}

%% file: sec/related_work.tex
\section{Related Work}

\paragraph{Reasoning Paths and Multi-trajectory Inference.}
Chain-of-thought prompting improves language-model reasoning by eliciting intermediate steps before the final answer~\citep{wei2022chain,kojima2022large}. 
Beyond single-path generation, self-consistency samples multiple reasoning paths for the same question and aggregates their answers~\citep{wang2022self}, while tree- and graph-based methods cast reasoning as search over candidate thoughts or states~\citep{yao2023tree,besta2024graph}. 
Other work evaluates or supervises intermediate reasoning through process supervision and step-level verification~\citep{lightman2023lets}. 
These methods exploit path structure to improve answer selection or reasoning performance; in contrast, we use multiple trajectories for the same question to study how reasoning quality is represented in hidden states.

\paragraph{Representation Analysis of Reasoning Trajectories.}
Recent work analyzes model behavior through internal representations, including latent knowledge, truthfulness, confidence, and task-relevant structure~\citep{burns2212discovering,azaria2023internal}, as well as representation geometry and activation directions for characterizing or intervening on behavior~\citep{li2023inference,park2023linear}. 
In reasoning tasks, intermediate hidden states have been shown to encode signals related to final-answer correctness, self-verification, and reasoning progress before the answer is generated~\citep{zhang2025reasoningmodelsknowtheyre,liu2025llm}. 
Most closely related to our work, \citet{sun2026llm} view reasoning as a structured trajectory in representation space, showing that activations near explicit reasoning markers form step-specific subspaces and that correct and incorrect trajectories may diverge over generation. 
Our work shares this trajectory-level view but addresses a different comparison problem: absolute step indices are not always comparable across trajectories, since the second step of one solution may be problem setup while the second step of another may already contain the main computation. 
We therefore align variable-length trajectories by relative phase and compare states within the same question and phase.

\paragraph{Correctness Signals and Activation Steering.}
Although correctness-related information can often be decoded from intermediate hidden states~\citep{zhang2025reasoningmodelsknowtheyre}, such signals are hard to interpret when examples are pooled across heterogeneous questions. 
A predictor trained across questions may capture question-level information, such as difficulty or prior solvability, rather than the quality of a particular reasoning trajectory. 
PAIR instead compares successful and unsuccessful trajectories within the same question and aligned phase, avoiding across-question correctness predictors. 
For causal validation, we apply the resulting phase-specific directions through activation steering. 
Prior work shows that hidden-state directions can steer model behavior at inference time~\citep{turner2023activation}, and contrastive activation methods further learn behavior-changing directions from paired examples~\citep{panickssery2023steering}.

%% file: sec/method.tex
\section{PAIR: Phase-Aligned Intra-question Reasoning}
\label{sec:method}

In this section, we introduce PAIR, a phase-aligned intra-question framework for analyzing and steering reasoning trajectories. 
PAIR is built on the premise that reasoning paths should not be compared as absolute-step sequences across heterogeneous questions. 
Instead, paths should be compared within the same question and at comparable relative phases of generation.

\begin{figure}[t]
    \centering
    \includegraphics[width=\linewidth]{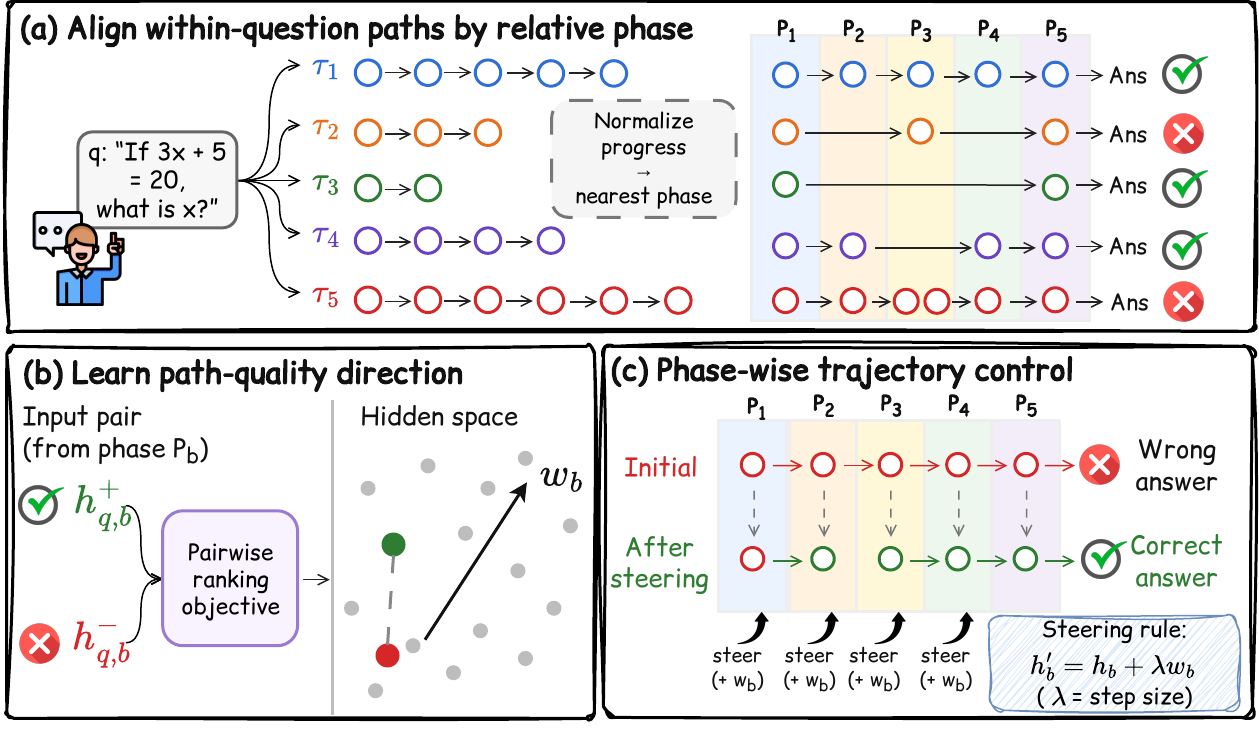}
    \caption{
Overview of PAIR.
(a) For each question, PAIR samples multiple reasoning trajectories and aligns variable-length paths into a fixed set of relative phase slots.
(b) Within each phase, PAIR constructs successful--unsuccessful trajectory pairs from the same question and learns a phase-specific path-quality direction using a pairwise ranking objective.
(c) The learned directions are used for phase-wise activation steering, where hidden states are shifted toward successful trajectory directions during generation.
    }
    \label{fig:framework}
    \vspace{-10pt}
\end{figure}

As shown in Fig.~\ref{fig:framework}, PAIR consists of three stages. 
First, for each question, we sample multiple reasoning trajectories and align their hidden states into a fixed set of relative phase slots. 
Second, within each phase, we construct successful--unsuccessful trajectory pairs from the same question and learn a phase-specific path-quality direction through a pairwise ranking objective. 
Third, we use the learned directions to steer hidden states during generation, thereby testing whether phase-specific path-quality signals can causally redirect reasoning trajectories.

\subsection{Problem Setup}
\label{sec:problem_setup}

Let $q$ denote a question with ground-truth answer $a_q^\ast$. 
For each question, we sample multiple reasoning trajectories from the model. 
The $i$-th trajectory is written as
\begin{equation}
\tau_{q,i}
=
\left(
x_{q,i,1:T_{q,i}},
H_{q,i,1:T_{q,i}},
\hat a_{q,i}
\right),
\label{eq:trajectory_definition}
\end{equation}
where $x_{q,i,1:T_{q,i}}$ are generated tokens, $H_{q,i,1:T_{q,i}}$ are token-level hidden states, and $\hat a_{q,i}$ is the final answer. 
The trajectory-level outcome is
\begin{equation}
y_{q,i}
=
\mathbb{I}\left[\hat a_{q,i}=a_q^\ast\right].
\label{eq:trajectory_outcome}
\end{equation}

As discussed in Sec.~\ref{sec:intro}, directly using $y_{q,i}$ as a label for every intermediate state can conflate local reasoning quality with final-answer correctness and question-level variation. 
Our goal is therefore not to predict final correctness from isolated hidden states. 
Instead, we ask whether one trajectory state is better than another when the question and the reasoning phase are matched.

After phase alignment, let $h_{q,i,b}^{(\ell)}$ denote the layer-$\ell$ representation of trajectory $\tau_{q,i}$ at phase $b$. 
For two trajectories sampled from the same question, we define the phase-wise path-quality preference:
\begin{equation}
h_{q,i,b}^{(\ell)} \succ h_{q,j,b}^{(\ell)}
\quad
\text{if}
\quad
y_{q,i}=1,\; y_{q,j}=0.
\label{eq:phase_wise_preference}
\end{equation}
Thus, final correctness is used to form preferences between trajectories matched by question and phase, rather than to label individual hidden states.

\subsection{Phase Alignment of Reasoning Trajectories}
\label{sec:phase_alignment}

PAIR first converts sampled reasoning trajectories into phase-aligned step representations.
This stage has two components: extracting step-level features from each trajectory and mapping variable-length step sequences to shared relative phases.

\paragraph{Step-level trajectory representation.}
For each question $q$, we sample $m$ reasoning trajectories under the same prompting and decoding configuration.
The trajectory and outcome notation follows Eqs.~\ref{eq:trajectory_definition}--\ref{eq:trajectory_outcome}.
Following \citet{sun2026llm}, we extract step-level representations using explicit reasoning markers.
We first detect step markers, such as ``Step 1'', numbered-list markers, and discourse markers such as ``First''; the full marker set is given in Appendix~\ref{app:step_marker}.
These markers indicate the beginning of reasoning steps.

Let $p_{q,i,k}$ denote the token position of the $k$-th detected step marker in trajectory $\tau_{q,i}$, and let $c_{q,i}$ denote the position of the conclusion marker.
If a trajectory contains $L_{q,i}$ reasoning steps, we represent step $j$ using the $r$ tokens immediately before the next step marker; the final step is represented using the $r$ tokens before the conclusion marker:
\begin{equation}
h_{q,i,j}^{(\ell)}
=
\begin{cases}
\operatorname{Pool}\!\left(
\left\{
H_{q,i,t}^{(\ell)}
:
p_{q,i,j+1}-r \leq t < p_{q,i,j+1}
\right\}
\right),
& j < L_{q,i}, \\[4pt]
\operatorname{Pool}\!\left(
\left\{
H_{q,i,t}^{(\ell)}
:
c_{q,i}-r \leq t < c_{q,i}
\right\}
\right),
& j = L_{q,i},
\end{cases}
\label{eq:step_representation}
\end{equation}
where $H_{q,i,t}^{(\ell)}$ is the token-level hidden state at layer $\ell$, and $\operatorname{Pool}(\cdot)$ denotes window aggregation.
This gives each trajectory a step-level hidden-state sequence
$\{h_{q,i,j}^{(\ell)}\}_{j=1}^{L_{q,i}}$.

\paragraph{Relative phase alignment.}
Since trajectories may contain different numbers of reasoning steps, absolute step indices are not directly comparable.
PAIR maps each trajectory to $K$ relative phase slots.
We set $K$ to the mode of the step counts in the training trajectories:
\begin{equation}
K
=
\operatorname{mode}
\left(
\left\{
L_{q,i}
:
q\in\mathcal{D}_{\mathrm{train}},
\;
i=1,\ldots,m
\right\}
\right).
\label{eq:num_phase_slots}
\end{equation}

For a trajectory with $L_{q,i}$ steps, step $j$ is assigned to its nearest relative phase:
\begin{equation}
b_{q,i,j}
=
1+
\operatorname{round}
\left(
\frac{(j-1)(K-1)}
{\max(L_{q,i}-1,1)}
\right),
\quad
b_{q,i,j}\in\{1,\ldots,K\}.
\label{eq:phase_assignment}
\end{equation}
This maps the first step to $P_1$ and the final step to $P_K$, while intermediate steps are placed according to normalized progress.
As illustrated in Fig.~\ref{fig:framework}(a), shorter trajectories may skip some phase slots, whereas longer trajectories may assign multiple steps to the same phase.

For each trajectory and phase, we collect the step representations assigned to that phase:
\begin{equation}
\mathcal{H}_{q,i,b}^{(\ell)}
=
\left\{
h_{q,i,j}^{(\ell)}
:
b_{q,i,j}=b
\right\}.
\label{eq:phase_hidden_set}
\end{equation}
The resulting phase-indexed sets are used to construct within-question comparisons in the next stage.

%
%
\subsection{Learning Phase-specific Path-quality Directions}
\label{sec:path_quality_direction}

After phase alignment, PAIR learns one path-quality direction for each phase.

For phase $b$, let $\mathcal{H}_{q,i,b}^{(\ell)}$ be the set of step representations from trajectory $\tau_{q,i}$ assigned to phase $b$, as defined in Eq.~\ref{eq:phase_hidden_set}.
We collect successful and unsuccessful phase-$b$ states for question $q$:
\begin{equation}
\mathcal{H}_{q,b}^{+}
=
\bigcup_{i:\,y_{q,i}=1}
\mathcal{H}_{q,i,b}^{(\ell)},
\qquad
\mathcal{H}_{q,b}^{-}
=
\bigcup_{i:\,y_{q,i}=0}
\mathcal{H}_{q,i,b}^{(\ell)}.
\label{eq:positive_negative_phase_sets}
\end{equation}

For each question $q$, PAIR constructs within-question preference pairs:
\begin{equation}
\mathcal{P}_{q,b}
=
\left\{
(h^{+},h^{-})
:
h^{+}\in \mathcal{H}_{q,b}^{+},
\;
h^{-}\in \mathcal{H}_{q,b}^{-}
\right\}.
\label{eq:within_question_pairs}
\end{equation}
Each pair states that, for the same question and same phase, the state from a successful trajectory should receive a higher path-quality score than the state from an unsuccessful trajectory.

For each phase $b$, we fit a linear scorer after phase-specific preprocessing.
Let $\phi_b(\cdot)$ denote the preprocessing map fitted on the training states for phase $b$.
The phase-$b$ score is
\begin{equation}
s_b(h)
=
\tilde w_b^\top \phi_b(h) + \beta_b,
\label{eq:phase_quality_score}
\end{equation}
where $\tilde w_b$ is the learned direction in the preprocessed space.
PAIR optimizes the pairwise logistic ranking objective:
\begin{equation}
\mathcal{L}_b
=
-\sum_{q}
\sum_{(h^{+},h^{-})\in \mathcal{P}_{q,b}}
\log
\sigma
\left(
s_b(h^{+}) - s_b(h^{-})
\right)
+
\lambda \|\tilde w_b\|_2^2.
\label{eq:pairwise_ranking_loss}
\end{equation}

Optimizing Eq.~\ref{eq:pairwise_ranking_loss} yields a phase-specific path-quality direction, which we map back to the original hidden-state space and normalize as $w_b$ for analysis and steering.
%
%
\subsection{Phase-wise Trajectory Steering}
\label{sec:phase_steering}

The learned directions are used as causal interventions on the reasoning trajectory.
As illustrated in Fig.~\ref{fig:framework}(c), PAIR modifies the hidden state according to the direction associated with the current reasoning phase.

Let $w_b$ be the normalized raw-space path-quality direction for phase $b$.
During generation, let $H_t^{(\ell)}$ denote the hidden state at layer $\ell$ and decoding position $t$.
Let $b_t \in \{1,\ldots,K\}$ be the phase assigned to the current position by the steering schedule.
PAIR applies the intervention
\begin{equation}
\widetilde{H}_t^{(\ell)}
=
H_t^{(\ell)}
+
\lambda_t w_{b_t},
\label{eq:phase_steering_update}
\end{equation}
where $\lambda_t$ is the intervention strength at position $t$.
The modified state $\widetilde{H}_t^{(\ell)}$ is passed to the remaining layers, with all model parameters fixed.

Rather than using a constant strength, we use the phase-specific score to adapt the intervention magnitude.
Let $\widetilde{s}_b(h)$ denote the learned phase-$b$ path-quality score, and let $\tau_b$ be a reference score estimated from successful training trajectories at phase $b$.
The intervention strength is
\begin{equation}
\lambda_t
=
\min
\left\{
\lambda_{\max},
\;
\alpha
\left[
\tau_{b_t}
-
\widetilde{s}_{b_t}
\left(
H_t^{(\ell)}
\right)
\right]_+
\right\},
\label{eq:adaptive_steering_strength}
\end{equation}
where $[x]_+=\max(x,0)$, $\alpha$ controls the global steering scale, and $\lambda_{\max}$ clips overly large updates.
Thus, PAIR pushes a state only when its phase-specific score falls below the reference level, and the clipping term prevents unstable interventions.

We evaluate two intervention schedules.
In single-phase steering, Eq.~\ref{eq:phase_steering_update} is applied only at one selected phase.
In progressive steering, the intervention follows the trajectory phase and uses the corresponding direction $w_{b_t}$ as generation advances.
The threshold construction, clipping value, and phase-schedule implementation are given in Appendix~\ref{app:steering_details}.

%% file: sec/exp.tex
\section{Experiments and Results}
\label{sec:results}

We structure our experiments to test the main claim that reasoning trajectories require phase-aligned within-question comparison. 
We first diagnose the limitations of standard across-question correctness probing, showing that it can conflate path quality with question-level variation and step misalignment (Sec.~\ref{sec:diagnosis}). 
We then evaluate whether PAIR learns phase-specific path-quality directions that improve within-question ranking of successful and unsuccessful trajectories (Sec.~\ref{sec:effectiveness}). 
Finally, we use phase-wise activation steering to test whether these directions can causally redirect generation rather than merely correlate with final correctness (Sec.~\ref{sec:steering}).
\subsection{Experimental Setup}
\label{sec:experimental_setup}
\paragraph{Models and benchmarks.}
We evaluate on three reasoning benchmarks: GSM8K~\cite{cobbe2021gsm8k}, MATH500~\cite{lightman2023lets}, and the logical\_deduction\_three\_objects task from Big-Bench Hard (BBH)~\cite{suzgun2022challenging}.
We use four open-weight models: LLaMA-3.1-8B-Instruct~\cite{meta_llama3_2024}, Qwen3-4B~\cite{qwen3technicalreport}, Gemma3-4B~\cite{gemma_2025}, and DeepSeek-R1-Distill-Llama-8B~\cite{deepseekai2025deepseekr1incentivizingreasoningcapability}.
For GSM8K, we train probes on 1{,}000 questions from the official training split and evaluate on 500 questions from the official test split.
For MATH and BBH, we use a fixed 70/30 question-level split, sampled once and reused across all methods and models.
All evaluations are conducted on held-out test questions, so trajectories from the same question never appear in both training and test sets.

\paragraph{Trajectory sampling and labeling.}
For each question, we sample 32 reasoning trajectories from the evaluated model with temperature $1.0$ and top-$p=0.95$.
A trajectory is labeled correct if its final extracted answer matches the ground truth, using dataset-specific extraction rules; unparsable outputs are treated as incorrect.
Sampling multiple trajectories per question lets us compare successful and unsuccessful paths under a shared question context, controlling for question-level difficulty.

\paragraph{Representation extraction.}
Unless otherwise specified, we probe the final transformer layer, where prior work has found the strongest signals for truthfulness and task outcomes~\citep{azaria2023internal,burns2212discovering,li2023inference}.
For each trajectory, we locate step markers and average the hidden states of the four tokens immediately preceding each marker to form a step-level representation.
These step-level representations are shared across all hidden-state scoring methods we compare, including final-token probing, pooled correctness probing, absolute-step alignment, normalized-step alignment, and our phase-aligned PAIR method.

\subsection{Diagnosing Ambiguities in Pointwise Correctness Analysis}
\label{sec:diagnosis}

\input{tabs/diagnose}

We first test whether pointwise correctness prediction remains reliable after controlling for question identity.
For each sampled trajectory, we extract the hidden states from a four-token window around the position closest to \(15\%\) relative progress and use their mean as the representation.
We then train a linear predictor using the trajectory-level correctness label \(y_{q,i}\).
Additional training details are given in Appendix~\ref{app:diagnostic_probe_details}.

We report the standard across-question evaluation (AQ), which tests whether the predictor can predict final correctness on held-out questions.
We contrast it with within-question evaluation (WQ), which ranks correct and incorrect trajectories sampled for the same question:
\begin{equation}\small
\mathrm{WQ\text{-}AUC}
=
\frac{1}{|\mathcal{Q}_{\pm}|}
\sum_{q \in \mathcal{Q}_{\pm}}
\frac{1}{|\mathcal{P}_q||\mathcal{N}_q|}
\sum_{i \in \mathcal{P}_q}
\sum_{j \in \mathcal{N}_q}
\mathbb{I}
\left[
s(x_{q,i}^{+}) > s(x_{q,j}^{-})
\right],
\label{eq:wq_auc}
\end{equation}
where $\mathcal{P}_q$ and $\mathcal{N}_q$ denote the successful and unsuccessful trajectories for question $q$, respectively, and $\mathcal{Q}_{\pm}$ contains questions with at least one trajectory from each group.
Balanced accuracy is computed under the same within-question protocol.

\begin{wrapfigure}{r}{0.5\linewidth}
    \vspace{-18pt}
    \centering
    \includegraphics[width=\linewidth]{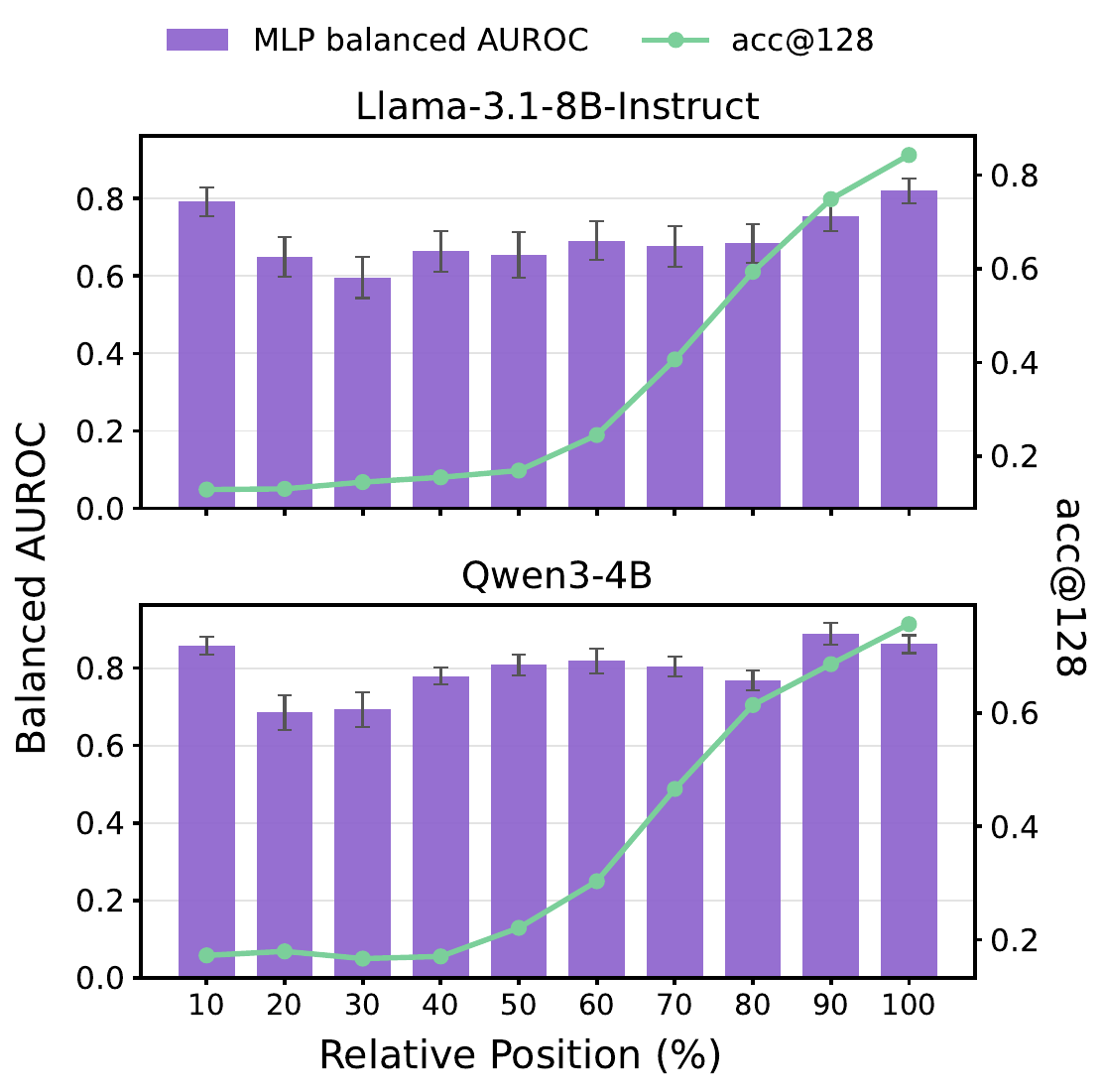}
\caption{
Balanced AUROC of correctness prediction and forced-answer accuracy across relative trajectory positions.
}
    \label{fig:relative_position_auc_acc}
    \vspace{-18pt}
\end{wrapfigure}

Table~\ref{tab:controlled_probe} shows that AQ performance consistently overestimates the strength of correctness signals.
Across models and benchmarks, AUROC and balanced accuracy drop sharply under WQ evaluation.
For example, the average BBH AUROC decreases from \(0.72\) to \(0.52\), and balanced accuracy decreases from \(0.70\) to \(0.50\).
This gap indicates that across-question correctness predictors capture substantial question-level information, such as difficulty or solvability, rather than only trajectory-level progress.

To test whether early correctness signals reflect answerability, we compare correctness prediction with forced-answer accuracy along the trajectory.
Following prior work on early correctness signals~\citep{zhang2025reasoningmodelsknowtheyre}, we train a two-layer MLP predictor on window-aggregated hidden states at different relative positions, using final-answer correctness as the label. Following the forced-answer extraction protocol of \citet{boppana2026reasoning}, we estimate answerability by truncating each trajectory at the same relative position, appending ``The answer is'', and sampling \(k=128\) completions with temperature \(0.8\).
We define \(\mathrm{acc@}k=c/k\), where \(c\) is the number of completions that produce the correct answer.

Fig.~\ref{fig:relative_position_auc_acc} shows the results on GSM8K for Llama-3.1-8B-Instruct and Qwen3-4B.
The MLP predictor achieves high balanced AUROC at very early positions, consistent with prior findings, while \(\mathrm{acc@}128\) remains close to zero.
This indicates that early representations can predict final correctness even when the prefix is not yet answerable.
Across relative positions, AUROC first decreases and then increases, whereas \(\mathrm{acc@}128\) rises as the trajectory approaches the final answer.
The late increase in AUROC is expected because later prefixes contain more information needed to recover the answer.
\emph{The early decrease may reflect a reduced influence of question-level information as generation moves away from the prompt.}
Together with Table~\ref{tab:controlled_probe}, this result supports the need to control question identity when analyzing pointwise correctness signals.

\subsection{Effectiveness of Phase-Aligned Path-quality Directions}
\label{sec:effectiveness}

\input{tabs/main_results}

\begin{figure}[t]
    \centering
    \phantomcaption
    \label{fig:phase_alignment_analysis}
    \setcounter{subfigure}{0}

    \captionsetup[subfigure]{
        labelformat=figsub,
        labelsep=period,
        font=small,
        justification=centering
    }
    
    \begin{subfigure}{0.35\linewidth}
        \centering
        
        \includegraphics[width=\linewidth]{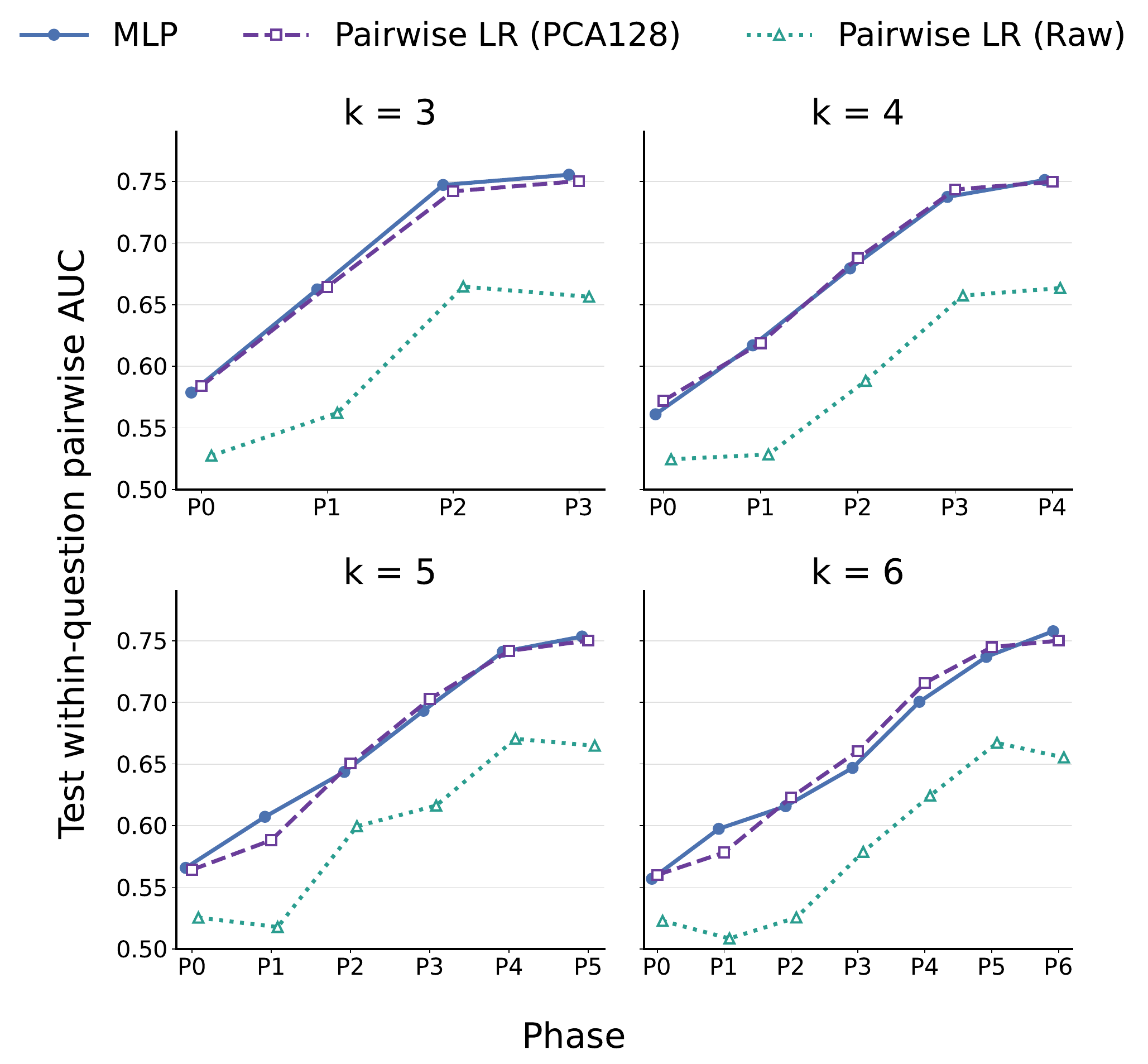}
        \caption{WQ-AUC across phases.}
        \label{fig:phase_auc}
    \end{subfigure}
    \hfill
    \begin{subfigure}{0.59\linewidth}
        \centering
        \includegraphics[width=\linewidth]{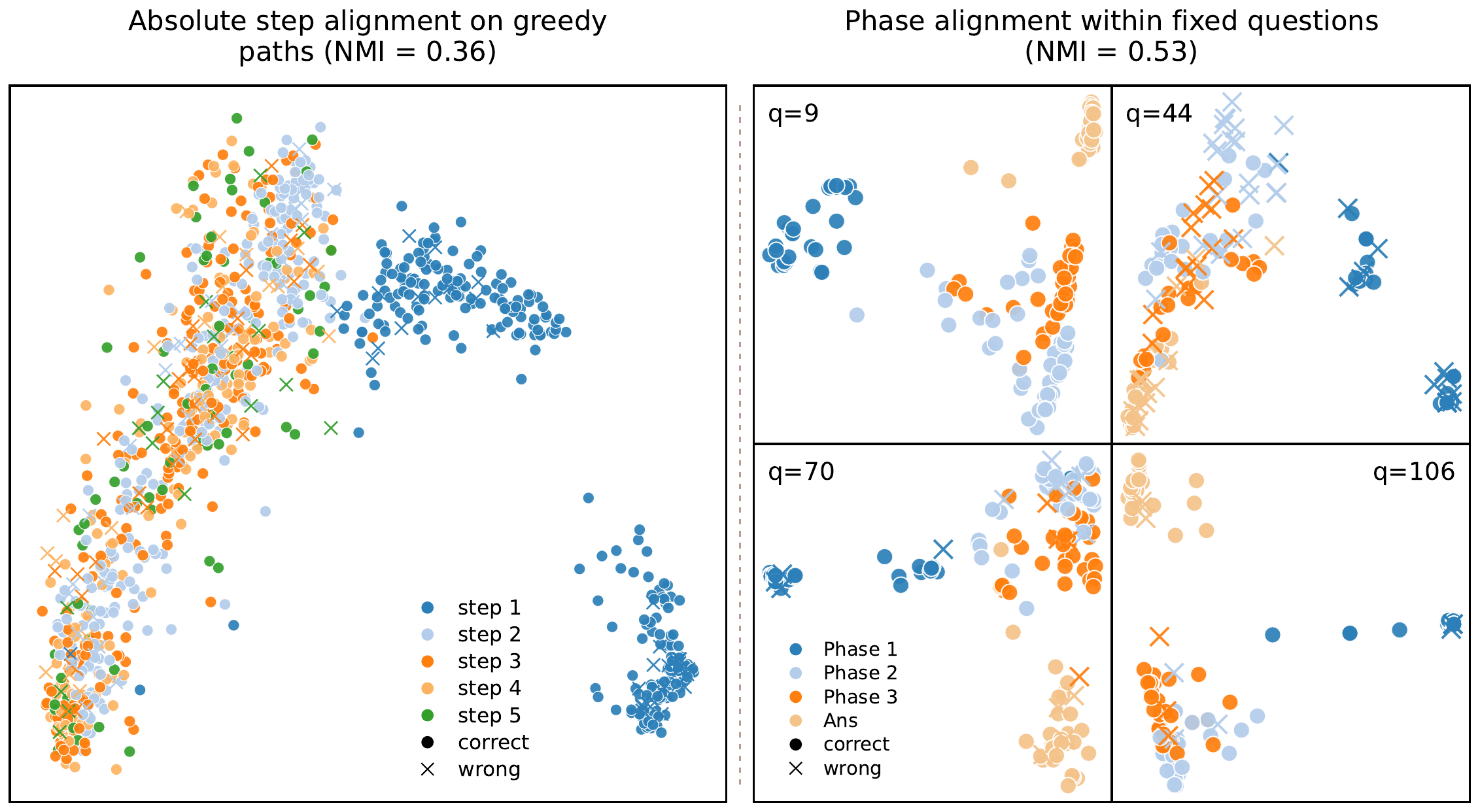}
        \caption{Absolute-step versus phase alignment.}
        \label{fig:cluster}
    \end{subfigure}
    \vspace{-10pt}
\end{figure}

We first examine how path-quality separability changes across aligned phases.
For each phase, we train a phase-specific scorer and evaluate it by pairwise AUROC (Eq.~\ref{eq:wq_auc}) on held-out questions, using \(K\in\{3,4,5,6\}\) phase bins.
As shown in Fig.~\ref{fig:phase_auc}, AUROC generally increases in later phases, indicating that later states better separate successful and unsuccessful trajectories.
This is consistent with prior findings that correct and incorrect trajectories diverge over generation~\citep{sun2026llm}.
Because the final phase gives the strongest and most stable separation, Table~\ref{tab:main_results} uses the final-phase scorer for trajectory selection.
We also find that raw pairwise LR is weaker, while PCA-compressed LR approaches the MLP, suggesting that linear directions are effective after reducing high-dimensional nuisance variation.
We therefore use the PCA-based linear scorer as the main PAIR estimator.

Table~\ref{tab:main_results} evaluates trajectory scoring for within-question ranking and Best-of-\(N\) selection.
In BoN@32, the model samples 32 reasoning trajectories for each question, and the scorer selects the highest-scoring trajectory as the final output.
We compare PAIR with four pointwise correctness baselines trained on greedy trajectories with final-answer correctness labels.
Pre-Ans token uses the representation immediately before the answer as a late-stage reference.
Pooled correctness uses the mean four-token representation near \(25\%\) relative progress, following prior work on early correctness signals~\citep{zhang2025reasoningmodelsknowtheyre}.
Absolute-step uses the third-step representation, reflecting step-index-based comparison~\citep{sun2026llm}.
Normalized-step uses the step representation closest to \(80\%\) relative progress.
PAIR achieves the best or competitive WQ-AUC and BoN@32 across most settings, showing that phase-aligned within-question training provides a stronger trajectory-ranking signal than baselines.
Pre-Ans token is also strong in several cases, but it is taken very close to the final answer, where answer information is often already recoverable.
The relatively strong Normalized-step baseline further suggests that relative trajectory position matters, even without full phase-aligned training.

Figure~\ref{fig:cluster} visualizes the effect of phase alignment.
When hidden states are grouped by absolute step indices on greedy paths, only the first step forms a clear cluster, and later steps are mixed, with an average NMI of \(0.36\).
This reflects that the same absolute step can correspond to different reasoning roles across questions.
After aligning trajectories within each fixed question by relative phase, phase groups become more separated, increasing the average NMI to \(0.53\).

\subsection{Phase-Specific Steering Redirects Reasoning Trajectories}
\label{sec:steering}
\input{tabs/steering}

We use phase-wise activation steering as a causal validation of the learned path-quality directions. 
Given a generated trajectory, we intervene at the corresponding reasoning phase by adding the learned direction to the hidden state, following Eq.~\ref{eq:phase_steering_update}. 
Table~\ref{tab:phase_wise_steering_all} reports steering results on LLaMA-3.1-8B-Instruct across benchmarks, comparing no intervention, single-phase steering, and progressive PAIR steering.
Single \(P_b\) applies only the direction learned for phase \(P_b\), while PAIR applies phase-specific directions as the trajectory advances. 
We report final-answer accuracy, wrong-to-correct flips \(W{\to}C\), correct-to-wrong flips \(C{\to}W\), and generated-token length after steering.

The steering results show that PAIR can correct some originally unsuccessful trajectories.
On GSM8K, accuracy increases from \(79.3\) to \(84.6\), with \(38.7\%\) of originally wrong trajectories flipped to correct and only \(3.3\%\) of originally correct trajectories flipped to wrong. 
MATH shows a similar accuracy gain, from \(42.0\) to \(47.3\), while BBH shows a smaller net gain because the large \(W{\to}C\) rate is partly offset by a higher \(C{\to}W\) rate. 
The single-phase results also suggest that early phases are more causally sensitive: steering at \(P_1\) or \(P_2\) is often more effective than steering at later phases. 
This is consistent with Fig.~\ref{fig:phase_auc} because readout separability and causal influence measure different aspects of the trajectory.
Later phases are easier to classify because they are closer to the final outcome, whereas earlier interventions can affect a longer portion of the remaining reasoning process.

%% file: tabs/diagnose.tex
\begin{table*}[t]
\centering
\caption{
Across-question and within-question evaluation of correctness probes.
For each benchmark, we report across-question (AQ) and within-question (WQ) AUROC and balanced accuracy, together with $\Delta=\mathrm{WQ}-\mathrm{AQ}$.
AQ/WQ entries are formatted as \textcolor{gray}{AQ}/\textbf{WQ}.
}
\label{tab:controlled_probe}
\setlength{\tabcolsep}{3.2pt}
\renewcommand{\arraystretch}{1.15}
\resizebox{\linewidth}{!}{
\begin{tabular}{lcccccccccccccccc}
\toprule
\multirow{2}{*}{Model}
& \multicolumn{4}{c}{GSM8K~\citep{cobbe2021gsm8k}}
& \multicolumn{4}{c}{MATH~\citep{lightman2023lets}}
& \multicolumn{4}{c}{BBH~\citep{suzgun2022challenging}}
\\
\cmidrule(lr){2-5}
\cmidrule(lr){6-9}
\cmidrule(lr){10-13}
& AUC & $\Delta$ & Balanced Acc. & $\Delta$
& AUC & $\Delta$ & Balanced Acc. & $\Delta$
& AUC & $\Delta$ & Balanced Acc. & $\Delta$ \\
\midrule
LLaMA-3.1-8B-Instruct
& \textcolor{gray}{0.72}/\textbf{0.56} & \textcolor{red}{-0.16} & \textcolor{gray}{0.66}/\textbf{0.51} & \textcolor{red}{-0.15}
& \textcolor{gray}{0.65}/\textbf{0.49} & \textcolor{red}{-0.16} & \textcolor{gray}{0.64}/\textbf{0.50} & \textcolor{red}{-0.14}
& \textcolor{gray}{0.72}/\textbf{0.53} & \textcolor{red}{-0.19} & \textcolor{gray}{0.66}/\textbf{0.51} & \textcolor{red}{-0.15}
\\

Qwen3-4B
& \textcolor{gray}{0.72}/\textbf{0.53} & \textcolor{red}{-0.19} & \textcolor{gray}{0.73}/\textbf{0.64} & \textcolor{red}{-0.09}
& \textcolor{gray}{0.64}/\textbf{0.51} & \textcolor{red}{-0.13} & \textcolor{gray}{0.60}/\textbf{0.57} & \textcolor{red}{-0.03}
& \textcolor{gray}{0.77}/\textbf{0.53} & \textcolor{red}{-0.24} & \textcolor{gray}{0.76}/\textbf{0.51} & \textcolor{red}{-0.25}
\\

Gemma3-4B
& \textcolor{gray}{0.65}/\textbf{0.50} & \textcolor{red}{-0.15} & \textcolor{gray}{0.63}/\textbf{0.52} & \textcolor{red}{-0.11}
& \textcolor{gray}{0.68}/\textbf{0.54} & \textcolor{red}{-0.14} & \textcolor{gray}{0.61}/\textbf{0.58} & \textcolor{red}{-0.03}
& \textcolor{gray}{0.71}/\textbf{0.51} & \textcolor{red}{-0.20} & \textcolor{gray}{0.71}/\textbf{0.41} & \textcolor{red}{-0.30}
\\

DeepSeek-R1-Distill-Llama-8B
& \textcolor{gray}{0.63}/\textbf{0.62} & \textcolor{red}{-0.01} & \textcolor{gray}{0.62}/\textbf{0.59} & \textcolor{red}{-0.03}
& \textcolor{gray}{0.65}/\textbf{0.51} & \textcolor{red}{-0.14} & \textcolor{gray}{0.64}/\textbf{0.56} & \textcolor{red}{-0.08}
& \textcolor{gray}{0.68}/\textbf{0.51} & \textcolor{red}{-0.17} & \textcolor{gray}{0.66}/\textbf{0.57} & \textcolor{red}{-0.09}
\\

\midrule
\rowcolor{gray!20}Average
& \textcolor{gray}{0.68}/\textbf{0.55} & \textcolor{red}{-0.13} & \textcolor{gray}{0.66}/\textbf{0.57} & \textcolor{red}{-0.10}
& \textcolor{gray}{0.66}/\textbf{0.51} & \textcolor{red}{-0.14} & \textcolor{gray}{0.62}/\textbf{0.55} & \textcolor{red}{-0.07}
& \textcolor{gray}{0.72}/\textbf{0.52} & \textcolor{red}{-0.20} & \textcolor{gray}{0.70}/\textbf{0.50} & \textcolor{red}{-0.20}
\\
\bottomrule
\end{tabular}
}
\vspace{-10pt}
\end{table*}

%% file: tabs/main_results.tex
\renewcommand{\arraystretch}{0.9}
\begin{table*}[t]
\centering
\caption{Evaluation of trajectory-quality estimation and Best-of-32 trajectory selection.
We highlight the best value in \textbf{\colorbox{mygreen}{green}} \textbf{with bold text} and the second best value in \colorbox{myblue}{\underline{blue}}.
}
\label{tab:main_results}
\resizebox{0.9\linewidth}{!}{
\begin{tabular}{llcccccccc}
\toprule
\multirow{2}{*}{Model}
& \multirow{2}{*}{Method}
& \multicolumn{2}{c}{GSM8K~\citep{cobbe2021gsm8k}}
& \multicolumn{2}{c}{MATH~\citep{lightman2023lets}}
& \multicolumn{2}{c}{BBH~\citep{suzgun2022challenging}}\\
\cmidrule(lr){3-4}
\cmidrule(lr){5-6}
\cmidrule(lr){7-8}
&
& WQ-AUC $\uparrow$ & BoN@32 $\uparrow$
& WQ-AUC $\uparrow$ & BoN@32 $\uparrow$
& WQ-AUC $\uparrow$ & BoN@32 $\uparrow$ \\
\midrule

\multirow{5}{*}{\makecell[c]{LLaMA-3.1\\-8B-Instruct~\cite{meta_llama3_2024}}}
& Pre-Ans token
& \cellcolor{myblue}{\underline{0.71}} & 83.3
& \cellcolor{mygreen}{\textbf{0.83}} & \cellcolor{mygreen}{\textbf{46.6}}
& \cellcolor{myblue}{\underline{0.65}} & \cellcolor{myblue}{\underline{84.0}}\\
& Pooled correctness
& 0.66 & 84.6
& 0.69 & 36.0
& 0.51 & 77.3\\
& Absolute-step
& 0.63 & 81.6
& 0.63 & 36.6
& 0.57 & 81.3\\
& Normalized-step
& 0.67 & \cellcolor{myblue}{\underline{85.3}}
& 0.71 & \cellcolor{myblue}{\underline{40.0}}
& 0.58 & 68.0 \\
& PAIR
& \cellcolor{mygreen}{\textbf{0.76}} &\cellcolor{mygreen}{\textbf{87.3}}
& \cellcolor{myblue}{\underline{0.77}} & \cellcolor{mygreen}{\textbf{46.6}}
& \cellcolor{mygreen}{\textbf{0.74}} & \cellcolor{mygreen}{\textbf{92.0}}\\
\midrule

\multirow{5}{*}{Qwen3-4B~\citep{qwen3technicalreport}}
& Pre-Ans token
& 0.48 & 58.7
& \cellcolor{myblue}{\underline{0.79}} & \cellcolor{myblue}{\underline{83.3}}
& \cellcolor{myblue}{\underline{0.84}} & 93.3\\
& Pooled correctness
& 0.61 & \cellcolor{myblue}{\underline{74.7}}
& 0.54 & 74.6
& 0.65 & 92.0\\
& Absolute-step
& 0.59 & 55.3
& 0.52 & 73.3
& 0.61 & 85.0 \\
& Normalized-step
& \cellcolor{myblue}{\underline{0.68}} &\cellcolor{mygreen}{\textbf{80.7}}
& 0.60 & 76.0
& 0.62 & \cellcolor{myblue}{\underline{94.6}}\\
& PAIR
& \cellcolor{mygreen}{\textbf{0.91}} & \cellcolor{mygreen}{\textbf{80.7}}
& \cellcolor{mygreen}{\textbf{0.89}} & \cellcolor{mygreen}{\textbf{84.7}}
& \cellcolor{mygreen}{\textbf{0.86}} & \cellcolor{mygreen}{\textbf{97.3}}\\
\midrule

\multirow{5}{*}{Gemma3-4B~\citep{gemma_2025}}
& Pre-Ans token
& \cellcolor{myblue}{\underline{0.60}} & \cellcolor{myblue}{\underline{89.3}}
& \cellcolor{myblue}{\underline{0.60}} & \cellcolor{myblue}{\underline{68.7}}
& \cellcolor{myblue}{\underline{0.67}} & \cellcolor{myblue}{\underline{94.6}}\\
& Pooled correctness
& 0.48 & \cellcolor{myblue}{\underline{89.3}}
& 0.53 & 61.3
& 0.51 & 92.0\\
& Absolute-step
& 0.55 & 83.3
& 0.49 & 55.3
& 0.55 & \cellcolor{myblue}{\underline{94.6}} \\
& Normalized-step
& 0.57 & 88.0
& 0.54 & 66.7
& 0.62 & \cellcolor{mygreen}{\textbf{96.0}}\\
& PAIR
& \cellcolor{mygreen}{\textbf{0.61}} & \cellcolor{mygreen}{\textbf{95.0}}
& \cellcolor{mygreen}{\textbf{0.74}} & \cellcolor{mygreen}{\textbf{75.3}}
& \cellcolor{mygreen}{\textbf{0.75}} & \cellcolor{mygreen}{\textbf{96.0}}\\
\midrule

\multirow{5}{*}{\makecell[c]{DeepSeek-R1-\\Distill-Llama-8B~\cite{deepseekai2025deepseekr1incentivizingreasoningcapability}}}
& Pre-Ans token
& \cellcolor{myblue}{\underline{0.65}} & 86.0
& \cellcolor{mygreen}{\textbf{0.87}} & \cellcolor{myblue}{\underline{84.6}}
& \cellcolor{myblue}{\underline{0.92}} & \cellcolor{mygreen}{\textbf{100.0}}\\
& Pooled correctness
& 0.60 & \cellcolor{myblue}{\underline{88.0}}
& 0.53 & 72.0
& 0.56 & \cellcolor{myblue}{\underline{81.3}}\\
& Absolute-step
& 0.58 & 84.3
& 0.63 & 74.6
& 0.50 & 69.3\\
& Normalized-step
& 0.57 & 86.6
& \cellcolor{myblue}{\underline{0.66}} & 80.0
& 0.84 & \cellcolor{mygreen}{\textbf{100.0}} \\
& PAIR
& \cellcolor{mygreen}{\textbf{0.75}} & \cellcolor{mygreen}{\textbf{89.3}}
& \cellcolor{mygreen}{\textbf{0.87}} & \cellcolor{mygreen}{\textbf{86.7}}
& \cellcolor{mygreen}{\textbf{0.94}} & \cellcolor{mygreen}{\textbf{100.0}}\\

\bottomrule
\end{tabular}
}
\vspace{-10pt}
\end{table*}

%% file: tabs/steering.tex
\begin{table*}[t]
\centering
\caption{
Phase-wise steering results across benchmarks.
Baseline denotes generation without intervention.
Single \(P_b\) applies the phase-\(b\) direction only at phase \(P_b\), while PAIR applies phase-specific directions progressively along the trajectory.
We report final-answer accuracy, wrong-to-correct and correct-to-wrong flip rates, and generated-token length after steering.
}
\label{tab:phase_wise_steering_all}
\newcommand{\pmgray}[1]{\textcolor{gray}{\text{\scriptsize$\pm$#1}}}
\resizebox{0.9\linewidth}{!}{
\begin{tabular}{lcccccc cccccc cccccc cccccc}
\toprule
\multirow{2}{*}{Intervention}
& \multicolumn{4}{c}{GSM8K}
& \multicolumn{4}{c}{MATH}
& \multicolumn{4}{c}{BBH} \\
\cmidrule(lr){2-5}
\cmidrule(lr){6-9}
\cmidrule(lr){10-13}
& Acc.  & $W{\rightarrow}C$ & $C{\rightarrow}W$ & Tok.
& Acc.  & $W{\rightarrow}C$ & $C{\rightarrow}W$ & Tok.
& Acc.  & $W{\rightarrow}C$ & $C{\rightarrow}W$ & Tok.
 \\
\midrule
\midrule
Baseline
& 79.3 & -- & -- & 234\pmgray{70}
& 42.0 & -- & -- & 383\pmgray{129} 
& 74.6 & -- & -- & 146\pmgray{28} \\

\cmidrule(lr){2-13}

Single $P_1$
& 83.3 & 32.2 & 3.3 & 242\pmgray{71} 
& 47.3 & 12.6 & 4.7 & 399\pmgray{181} 
& 76.0 & 57.8 & 17.9 & 167\pmgray{25} \\

Single $P_2$
& 82.6 & 25.8 & 2.5 & 233\pmgray{67} 
& 44.7 & 6.9 & 3.2 & 386\pmgray{121}
& 70.6 & 5.2 & 7.1 & 187\pmgray{34} \\

Single $P_3$
& 80.0 &3.2 & 0 & 234\pmgray{70} 
& 42.0 & 2.3 & 3.2 & 387\pmgray{170} 
& 69.3 & 36.8 & 19.6 & 205\pmgray{66} \\

Single $P_4$
& 79.3 & 0.0 & 0.0 & 234\pmgray{70} 
& 42.7 & 1.1 & 0 & 386\pmgray{123} 
& 76.0 & 10.5 & 1.8 & 198\pmgray{54}\\

Single $P_5$
& 79.3 & 0.0 & 0.0 & 234\pmgray{70}
& 42.0 & 0.0 & 0.0 & 383\pmgray{129} 
& 72.0 & 0.0 & 3.6 & 169\pmgray{26} \\

\cmidrule(lr){2-13}

\rowcolor{gray!20} PAIR (Ours)
& 84.6 & 38.7 & 3.3 & 244\pmgray{74} 
& 47.3 & 12.6 & 4.7 & 398\pmgray{181}  
& 76.0 & 57.8 & 17.9 & 167\pmgray{25} \\

\bottomrule
\end{tabular}
}
\vspace{-10pt}
\end{table*}

%% file: sec/conclusion.tex
\section{Conclusion}
\label{sec:conclusion}

We presented a phase-aligned view of reasoning trajectories in LLMs. 
We showed that standard correctness probes can be ambiguous: across-question evaluation may capture question-level variation, and absolute step indices may compare different reasoning stages. 
PAIR addresses this by comparing successful and unsuccessful trajectories from the same question at matched phases to learn path-quality directions. 
Phase-wise steering further shows that these directions can change reasoning outcomes rather than only predict them. 
These results suggest analyzing reasoning trajectories as structured processes, controlling comparisons by both question identity and relative phase.

%% file: sec/appendix.tex
\clearpage
\appendix

\section{Appendix}
\subsection{Prompt Design}
\label{app:prompt_template}
The prompt for each model are listed as below:
\begin{figure}[h]
    \centering
    \includegraphics[width=\linewidth]{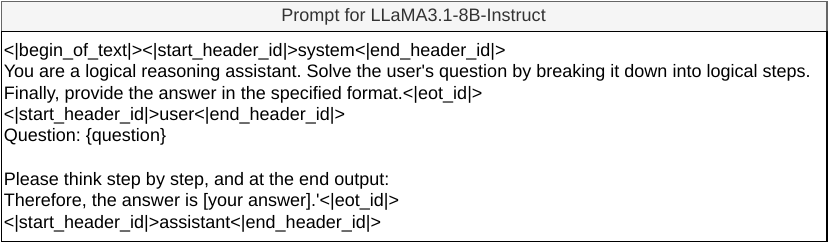}
    \caption{Prompt for LLaMA3.1-8B-Instruct}
\end{figure}
\begin{figure}[h]
    \centering
    \includegraphics[width=\linewidth]{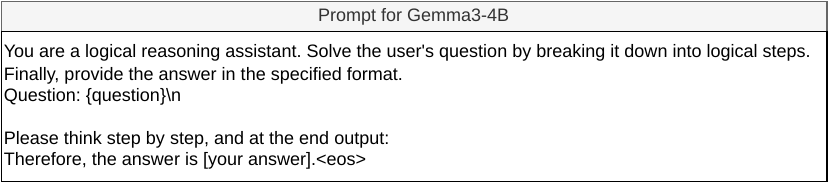}
    \caption{Prompt for Gemma3-4B}
\end{figure}
\begin{figure}[h]
    \centering
    \includegraphics[width=\linewidth]{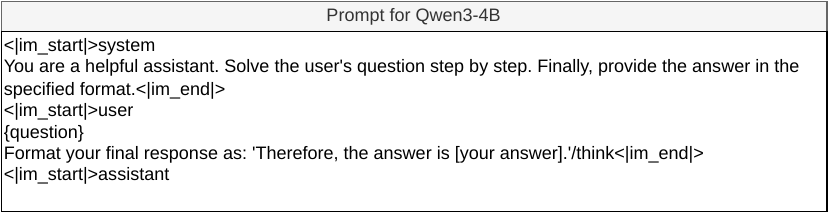}
    \caption{Prompt for Qwen3-4B}
\end{figure}
\begin{figure}[h]
    \centering
    \includegraphics[width=\linewidth]{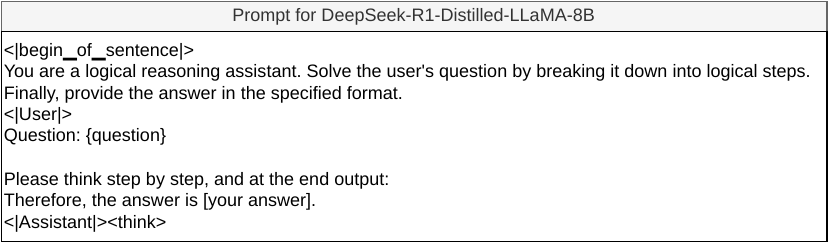}
    \caption{Prompt for DeepSeek-R1-Distilled-LLaMA-8B}
\end{figure}

\subsection{Model Details}
\label{app:model_details}

We evaluate four open-weight LLMs with different instruction-tuning and reasoning-training backgrounds. 
All experiments use the publicly released checkpoints without additional finetuning. 
Unless otherwise specified, hidden-state features are extracted from the final transformer layer, and steering interventions are applied to the same layer.

\subsubsection{Llama-3.1-8B-Instruct}
\label{app:model_llama}

Llama-3.1-8B-Instruct is an instruction-tuned model from the Llama 3.1 family~\citep{meta_llama3_2024}. 
We use it as the primary model for both diagnostic analysis and phase-wise steering because it provides stable multi-step reasoning trajectories under chain-of-thought prompting. 
For all experiments, we use the same decoding and trajectory-sampling settings described in Sec.~\ref{sec:experimental_setup}.

\subsubsection{Qwen3-4B}
\label{app:model_qwen}

Qwen3-4B is an open-weight model from the Qwen3 family~\citep{qwen3technicalreport}. 
We include it to test whether the observed trajectory-level patterns hold beyond the Llama model family. 
In our experiments, Qwen3-4B is evaluated with the same prompting format, trajectory-sampling configuration, and hidden-state extraction procedure as the other models.

\subsubsection{Gemma3-4B}
\label{app:model_gemma}

Gemma3-4B is an open-weight model from the Gemma 3 family~\citep{gemma_2025}. 
We use this model as an additional architecture family for evaluating pointwise correctness analysis, phase-aligned trajectory ranking, and Best-of-\(N\) trajectory selection. 
No model-specific finetuning or calibration is applied.

\subsubsection{DeepSeek-R1-Distill-Llama-8B}
\label{app:model_deepseek}

DeepSeek-R1-Distill-Llama-8B is a distilled reasoning model from the DeepSeek-R1 series~\citep{deepseekai2025deepseekr1incentivizingreasoningcapability}. 
We include it to evaluate PAIR on a model with stronger explicit reasoning behavior. 
The model is evaluated with the same trajectory-sampling and representation-extraction pipeline used for the other models. 
Because distilled reasoning models tend to generate longer reasoning traces, we report generated-token lengths when evaluating steering interventions.

\subsection{Dataset Details}
\label{app:dataset_details}

We evaluate on three reasoning benchmarks: GSM8K, MATH-500, and BBH. 
All splits are made at the question level. 
For every benchmark, trajectories sampled from the same question are assigned to the same split, so no question appears in both training and test sets.

\subsubsection{GSM8K}
\label{app:data_gsm8k}

GSM8K is a grade-school math word-problem benchmark~\citep{cobbe2021gsm8k}. 
We use GSM8K as the main benchmark for learning phase-specific path-quality directions because it provides a sufficiently large training split. 
Specifically, we train on 1,000 questions sampled from the official training split and evaluate on 500 questions sampled from the official test split. 
For each question, we sample 32 reasoning trajectories with temperature \(1.0\) and top-\(p=0.95\). 
A trajectory is labeled correct if the extracted final answer matches the ground-truth answer; otherwise, it is labeled incorrect.

\subsubsection{MATH-500}
\label{app:data_math}

MATH-500 is a 500-problem subset of the MATH benchmark~\citep{lightman2023lets}. 
It contains competition-style mathematical reasoning problems and is substantially more difficult than GSM8K. 
Since MATH-500 does not provide a large official training split for our setting, we use a fixed 70/30 question-level split. 
The split is sampled once and reused across all models and methods to ensure a consistent comparison. 
For each question, we sample 32 reasoning trajectories using the same decoding configuration as GSM8K, and correctness is determined by dataset-specific final-answer extraction.

\subsubsection{BBH Logical Deduction}
\label{app:data_bbh}

We use the \texttt{logical\_deduction\_three\_objects} task from Big-Bench Hard~\citep{suzgun2022challenging}. 
This task evaluates symbolic and logical reasoning over a small set of objects and constraints. 
As with MATH-500, we use a fixed 70/30 question-level split and reuse the same split across all methods and models. 
For each question, we sample 32 reasoning trajectories with temperature \(1.0\) and top-\(p=0.95\). 
A trajectory is labeled correct if its extracted final answer matches the ground-truth option or answer string under the task-specific extraction rule.

\subsection{Implementation Details}

\subsubsection{Step Marker Extraction}
\label{app:step_marker}

We identify intermediate reasoning steps from the model's generated response using a rule-based \emph{smart marker} procedure. The input is first normalized with the same text cleaning used elsewhere in our pipeline, and step markers are then detected on the cleaned response text.

\paragraph{Candidate step markers.}
The smart marker extractor considers four classes of candidate boundaries:
\begin{enumerate}
    \item \textbf{Explicit numeric markers at line start}, such as \texttt{Step 3:}, \texttt{3.}, \texttt{3)}, or \texttt{(3)}.
    \item \textbf{Section-style numeric headers} at line start, such as \texttt{Part 2:}, \texttt{Phase 3}, \texttt{Round 4}, etc.
    \item \textbf{Discourse markers}, such as \texttt{First}, \texttt{Second}, \texttt{Next}, \texttt{Then}, \texttt{After that}, \texttt{Finally}, and \texttt{Lastly}. These are only accepted when they occur at a line boundary or at a sentence boundary.
    \item \textbf{Heading-like lines} ending with a colon, provided that they look like short natural-language headers rather than equations or formatting artifacts.
\end{enumerate}

\paragraph{Filtering heuristics.}
To avoid spurious boundaries, we apply several filters. Numeric or discourse candidates are kept only if the following text looks like genuine prose rather than a formula, a bare number, or a final-answer statement. We explicitly reject separator/table lines (e.g., rows of dashes), equation-only lines, and lines that contain only symbols or numbers. We also strip common Markdown decoration before testing whether a line looks like a valid heading or step.

\paragraph{Preference for explicit structure.}
If the response already contains a clear explicit step structure (at least two numbered or \texttt{Step-$n$} markers), we trust that structure and ignore weaker discourse-style transitions inside those numbered steps. Otherwise, we merge numeric, discourse, and heading candidates into a single ordered candidate list.

\paragraph{Step numbering.}
Candidates are processed from left to right and assigned monotonically increasing step indices. Explicit numeric markers keep their stated index. Discourse ordinals such as \texttt{First}, \texttt{Second}, and \texttt{Third} are mapped to their canonical indices. Other accepted discourse or heading markers are assigned the next available step index. We discard any candidate whose assigned index would be non-monotonic or implausibly large.

\paragraph{Mapping markers to hidden states.}
After converting accepted character positions to token indices, we represent step $k$ using the hidden state immediately before the marker of step $k{+}1$. Thus, a marker for step $k{+}1$ defines the feature for the preceding step $k$. In addition, we extract:
\begin{itemize}
    \item a \textbf{final-step} feature from the hidden state immediately before the last conclusion marker (e.g., phrases such as \texttt{therefore}, \texttt{so the answer is}, or equivalent conclusion patterns), and
    \item an \textbf{answer-pre} feature from the hidden state immediately before the predicted answer token.
\end{itemize}

In our implementation, if an answer token can be located, we prefer the last conclusion marker that appears before that answer; otherwise we use the last detected conclusion marker in the response.

\subsubsection{Implementation details for learning path-quality directions.}
For each phase, we train an independent pairwise logistic ranking model.
The input feature is the mean hidden state over the four tokens immediately preceding the extracted boundary, corresponding to the \texttt{window4\_mean} feature.
The preprocessing map $\phi_b$ is fitted separately for each phase using only training questions.
It consists of standardization followed by PCA:
\begin{equation}
\phi_b(h)
=
P_b
\left(
\frac{h-\mu_b}{\sigma_b}
\right),
\label{eq:preprocessing_map}
\end{equation}
where $\mu_b$ and $\sigma_b$ are the training-set mean and standard deviation for phase $b$, and $P_b$ is the PCA projection matrix.
We use at most 128 PCA components, with the actual number of components set to
\begin{equation}
d_b
=
\min(128, N_b-1, D),
\label{eq:pca_dimension}
\end{equation}
where $N_b$ is the number of training states for phase $b$ and $D$ is the original hidden dimension.
If PCA is not applicable, $\phi_b$ reduces to standardization.

For each question and phase, we form all positive--negative pairs from $\mathcal{H}_{q,b}^{+}\times\mathcal{H}_{q,b}^{-}$.
To control the number of training examples, we keep at most 128 positive--negative pairs per question, sampled uniformly when more are available.
In the logistic-regression implementation, we add both the forward difference and the reversed difference:
\begin{equation}
x = \phi_b(h^{+})-\phi_b(h^{-}),\quad y=1,
\qquad
x' = \phi_b(h^{-})-\phi_b(h^{+}),\quad y'=0.
\label{eq:pairwise_difference_examples}
\end{equation}
This is equivalent to optimizing the ranking objective in Eq.~\ref{eq:pairwise_ranking_loss}.

We use \texttt{sklearn.linear\_model.LogisticRegression} with \texttt{solver=lbfgs}, \texttt{class\_weight=balanced}, $C=1.0$, and \texttt{max\_iter=2000}.
All splits are made at the question level: training, validation, and test questions are disjoint, and pairs are constructed only within the corresponding split.

For steering, the learned coefficient $\tilde w_b$ is mapped back to the raw hidden-state space.
When PCA is used as in Eq.~\ref{eq:preprocessing_map}, the raw-space direction is
\begin{equation}
w_b
=
\operatorname{diag}(\sigma_b)^{-1}
P_b^\top
\tilde w_b,
\qquad
\bar w_b
=
\frac{w_b}{\|w_b\|_2}.
\label{eq:raw_direction_mapping}
\end{equation}
The normalized direction $\bar w_b$ is used for phase-wise steering.

\subsubsection{Details for PAIR Direction Learning and Steering}
\label{app:steering_details}

\paragraph{Feature Preprocessing and Linear Scorer}
\label{app:direction_learning_details}

For each phase $b$, we train an independent linear path-quality scorer using only training questions.
The raw feature $h$ is the mean hidden state over the four tokens immediately preceding the extracted reasoning boundary:
\begin{equation}
h
=
\frac{1}{4}
\sum_{r=1}^{4}
H_{t-r}^{(\ell)} ,
\label{eq:appendix_window4_feature}
\end{equation}
where $H_{t-r}^{(\ell)}$ is the token-level hidden state at layer $\ell$.

Each phase has its own preprocessing map. 
We first standardize raw features using the training-set mean and standard deviation:
\begin{equation}
\bar h
=
\frac{h-\mu_b}{\sigma_b}.
\label{eq:appendix_standardization}
\end{equation}
We then apply PCA when the number of available training states is sufficient:
\begin{equation}
\phi_b(h)
=
P_b \bar h,
\label{eq:appendix_pca_map}
\end{equation}
where $P_b$ is the phase-specific PCA projection matrix. 
The PCA dimension is set to
\begin{equation}
d_b
=
\min(128, N_b-1, D),
\label{eq:appendix_pca_dim}
\end{equation}
where $N_b$ is the number of training states at phase $b$ and $D$ is the raw hidden dimension. 
If PCA is not applicable, $\phi_b$ reduces to standardization.

The phase-$b$ scorer is
\begin{equation}
\widetilde{s}_b(h)
=
\tilde w_b^\top \phi_b(h) + \beta_b,
\label{eq:appendix_phase_scorer}
\end{equation}
where $\tilde w_b$ and $\beta_b$ are learned by pairwise logistic regression.

For each question and phase, we form all successful--unsuccessful pairs
$(h^+,h^-)$ from the same question. 
To implement the pairwise objective, we add both the forward and reversed differences:
\begin{equation}
x
=
\phi_b(h^+) - \phi_b(h^-),
\quad
y=1,
\qquad
x'
=
\phi_b(h^-) - \phi_b(h^+),
\quad
y'=0.
\label{eq:appendix_pairwise_diffs}
\end{equation}
We keep at most 128 positive--negative pairs per question and phase, sampled uniformly when more pairs are available.

The logistic-regression model uses \texttt{lbfgs}, \texttt{class\_weight=balanced}, $C=1.0$, \texttt{max\_iter=2000}, and random seed 42.
All train, validation, and test splits are made at the question level.

\paragraph{Mapping Directions Back to Raw Hidden Space}
\label{app:raw_direction_mapping}

The learned coefficient $\tilde w_b$ lies in the preprocessed feature space.
For steering, we map it back to the original hidden-state space.
When preprocessing consists of standardization followed by PCA, the raw-space direction is
\begin{equation}
v_b
=
\operatorname{diag}(\sigma_b)^{-1} P_b^\top \tilde w_b .
\label{eq:appendix_raw_direction}
\end{equation}
We then normalize it:
\begin{equation}
w_b
=
\frac{v_b}{\|v_b\|_2}.
\label{eq:appendix_normalized_direction}
\end{equation}
This normalized raw-space direction $w_b$ is used for activation steering.

\paragraph{Thresholded and Clipped Phase-wise Steering}
\label{app:thresholded_steering}

For each phase $b$, we estimate a reference score from successful training trajectories.
In our implementation, we use the lower quartile of successful-state scores:
\begin{equation}
\tau_b
=
\operatorname{Quantile}_{0.25}
\left(
\left\{
\widetilde{s}_b(h)
:
h\in\mathcal{H}_{q,b}^{+},
\;
q\in\mathcal{D}_{\mathrm{train}}
\right\}
\right).
\label{eq:appendix_threshold}
\end{equation}
This threshold defines the lower reference range of successful trajectories at phase $b$.

During generation, let $H_t^{(\ell)}$ be the hidden state at layer $\ell$ and decoding position $t$.
Let $b_t$ be the phase assigned to the current position by the intervention schedule.
The adaptive steering strength is
\begin{equation}
\lambda_t
=
\min
\left\{
\lambda_{\max},
\;
\alpha
\left[
\tau_{b_t}
-
\widetilde{s}_{b_t}
\left(
H_t^{(\ell)}
\right)
\right]_+
\right\},
\label{eq:appendix_clipped_strength}
\end{equation}
where $[x]_+=\max(x,0)$, $\alpha$ is the global steering scale, and $\lambda_{\max}$ is the maximum allowed intervention magnitude.

The hidden-state update is
\begin{equation}
\widetilde{H}_t^{(\ell)}
=
H_t^{(\ell)}
+
\lambda_t w_{b_t}.
\label{eq:appendix_steering_update}
\end{equation}
The modified hidden state is passed to the remaining layers, while all model parameters remain fixed.

\paragraph{Intervention Schedules}
\label{app:intervention_schedules}

We evaluate two schedules.

\paragraph{Single-phase steering.}
For a selected phase $b$, the intervention is applied only when the current position belongs to that phase:
\begin{equation}
b_t=b
\quad\Rightarrow\quad
\widetilde{H}_t^{(\ell)}
=
H_t^{(\ell)}
+
\lambda_t w_b.
\label{eq:appendix_single_phase}
\end{equation}
No intervention is applied outside the selected phase.

\paragraph{Progressive steering.}
In progressive steering, the intervention follows the trajectory phase:
\begin{equation}
\widetilde{H}_t^{(\ell)}
=
H_t^{(\ell)}
+
\lambda_t w_{b_t},
\qquad
b_t\in\{1,\ldots,K\}.
\label{eq:appendix_progressive_phase}
\end{equation}
Thus, different parts of the generation are steered by the direction learned for their corresponding phase.
\paragraph{Steering Hyperparameters}
\label{app:steering_hyperparams}

For phase-wise steering, we use the adaptive intervention strength defined in Eq.~\ref{eq:adaptive_steering_strength}. 
The two main hyperparameters are the global scale \(\alpha\), which controls the size of the score-dependent update, and the clipping threshold \(\lambda_{\max}\), which limits the maximum intervention magnitude. 
We select these values by sweeping a small grid on the validation split and choosing the setting that gives the best validation accuracy while avoiding large correct-to-wrong flip rates. 
After selection, the same values are fixed and used for all test-set steering evaluations. 
Table~\ref{tab:steering_hyperparams} reports the hyperparameters used for the steering experiments in Table~\ref{tab:phase_wise_steering_all}.

\begin{table}[t]
\centering
\caption{
Steering hyperparameters selected by validation-set sweep.
\(\alpha\) controls the scale of the adaptive update, and \(\lambda_{\max}\) clips the maximum intervention magnitude.
}
\label{tab:steering_hyperparams}
\begin{tabular}{lcc}
\toprule
Benchmark & \(\alpha\) & \(\lambda_{\max}\) \\
\midrule
GSM8K & 5 & 5 \\
MATH-500 & 5 & 5 \\
BBH & 2.5 & 2.5 \\
\bottomrule
\end{tabular}
\end{table}

\paragraph{Pointwise Probe Training Details}
\label{app:diagnostic_probe_details}

In addition to the final pairwise probe, we train four \emph{single-path} probes. Each of these probes is trained as a binary classifier on \textbf{greedy} trajectories, where the label indicates whether the final answer of the trajectory is correct. The learned probe is then evaluated on \textbf{sampled} trajectories from the same dataset using within-question ranking metrics and best-of-$N$ selection. The four probes differ only in the hidden-state feature used as input.

\paragraph{Data sources.}
For each question, we use two kinds of trajectories:
\begin{itemize}
    \item a single \textbf{greedy} trajectory for probe training, and
    \item a set of \textbf{sampled} trajectories (typically $N=32$) for evaluation.
\end{itemize}
Greedy trajectories provide one labeled example per question for training. Sampled trajectories provide multiple correct and incorrect candidate paths for the same question, allowing us to evaluate whether a probe score can rank good reasoning paths above bad ones.

\paragraph{Outer train/test split.}
All splits are defined at the \textbf{question level}. Let $\mathcal{Q}$ denote the set of question IDs. We construct a fixed outer train/test split using only greedy-path correctness labels, with a default ratio of $70\%/30\%$. Thus, all greedy and sampled paths from the same question belong to the same outer split, preventing leakage across train and test.

\paragraph{Inner validation for hyperparameter selection.}
Within the outer training questions, we create an inner validation split used only to select the logistic regression regularization strength. When a dedicated validation split is not available, we create one by splitting the outer training questions again, typically using an inner validation ratio of $0.2$. The split is also performed at the question level. If stratified splitting is not possible due to small class counts, we fall back to an unstratified random split.

\paragraph{Feature representation.}
Each of the four probes takes as input a single hidden-state vector extracted from one location in the trajectory. In all cases, the hidden state is represented as the mean of the \textbf{four tokens immediately preceding} the target position. Let $h_{t-3}, h_{t-2}, h_{t-1}, h_t \in \mathbb{R}^d$ denote the hidden states of the four tokens ending at the target token position $t$. The feature vector is
\[
x = \frac{1}{4}\sum_{i=0}^{3} h_{t-i} \in \mathbb{R}^d.
\]

The four probes are defined as follows:
\begin{enumerate}
    \item \textbf{Answer-pre probe.}
    We locate the predicted answer token and use the mean hidden state of the four tokens immediately preceding that answer position.

    \item \textbf{Relative-position-25\% probe.}
    Let $L$ be the token length of the cleaned response. We compute the target token index as
    \[
    t_{0.25} = \mathrm{round}\!\left(0.25 \cdot (L-1)\right),
    \]
    and use the mean hidden state of the four-token window ending at $t_{0.25}$.

    \item \textbf{Step-3 probe.}
    We run the step marker extraction procedure described in Appendix~\ref{app:step_marker}. By our convention, the hidden state of step $k$ is taken from the hidden state immediately before the marker of step $k+1$. The step-3 probe therefore uses the four-token mean immediately before the marker of step 4. If a trajectory does not contain such a step boundary, it is skipped for this probe.

    \item \textbf{80\%-step probe.}
    We first extract the ordered list of reasoning-step features for the trajectory, including the final-step feature defined by the hidden state before the conclusion marker. If there are $m$ such reasoning-step features, we select the step at index
    \[
    j_{0.8} = \mathrm{round}\!\left(0.8 \cdot (m-1)\right),
    \]
    and use that step's four-token mean hidden state as the probe input.
\end{enumerate}

\paragraph{Training labels.}
For greedy training, each question contributes one trajectory with a binary label
\[
y \in \{0,1\},
\]
where $y=1$ indicates that the final predicted answer is correct and $y=0$ indicates that it is incorrect.

\paragraph{Preprocessing.}
For each probe, we collect all training feature vectors from the greedy training split. Let the raw feature matrix be
\[
X \in \mathbb{R}^{n \times d}.
\]
We first standardize features dimension-wise using a \texttt{StandardScaler}:
\[
\tilde{X}_{ij} = \frac{X_{ij} - \mu_j}{\sigma_j}.
\]
We then fit PCA on the standardized training features and project them to a low-dimensional space. If the requested PCA dimension is $k$, the actual number of retained components is
\[
k' = \min(k, n-1, d).
\]
In our experiments the default target dimension is $k=128$, so typically the probe input after PCA is
\[
z \in \mathbb{R}^{128}.
\]

\paragraph{Classifier.}
Each probe is a logistic regression classifier trained on the PCA-transformed features. For a transformed feature vector $z$, the probe predicts
\[
s(z) = w^\top z + b,
\]
and the probability of correctness is
\[
p(y=1 \mid z) = \sigma(w^\top z + b),
\]
where $\sigma(\cdot)$ is the logistic sigmoid.

We use \texttt{sklearn.linear\_model.LogisticRegression} with:
\begin{itemize}
    \item solver: \texttt{lbfgs},
    \item class weighting: \texttt{balanced},
    \item maximum iterations: \texttt{5000},
    \item random seed fixed for reproducibility.
\end{itemize}

\paragraph{Hyperparameter selection.}
The main tuned hyperparameter is the inverse regularization strength $C$ of logistic regression. We evaluate a candidate set such as
\[
C \in \{0.03, 0.1, 0.3, 1.0, 3.0\}.
\]
For each candidate $C$, we train on the greedy inner-training questions and evaluate on sampled paths from the inner-validation questions. Although the probe is trained on single-path correctness labels, model selection is based on \textbf{within-question ranking performance} on sampled validation paths, since ranking is the downstream behavior of interest.

Concretely, for each candidate $C$, we score sampled validation paths and compute:
\begin{itemize}
    \item pooled within-question AUC,
    \item path-level AUROC across sampled paths,
    \item and optionally best-of-$N$ accuracy.
\end{itemize}
We choose the candidate with the best validation pooled within-question AUC; ties are broken by path-level AUROC and then by preferring smaller $C$ (stronger regularization).

\paragraph{Refitting.}
After selecting $C^\star$, we refit the probe on all greedy trajectories from the outer training questions using the same preprocessing pipeline:
\begin{enumerate}
    \item fit \texttt{StandardScaler} on outer-train greedy features,
    \item fit PCA on standardized outer-train features,
    \item train logistic regression with $C^\star$ on the transformed features.
\end{enumerate}
The fitted scaler, PCA parameters, and logistic regression coefficients are saved for later analysis and reuse.

\subsection{Compute Resources}
\label{app:compute_resources}

The main computational cost of our experiments comes from trajectory sampling and hidden-state caching, rather than from training or steering. 
For each model and benchmark, we sample 32 trajectories per question and store the hidden states needed for later phase alignment and scoring. 
On a single NVIDIA A100 GPU, one full trajectory-sampling run for a model--benchmark setting takes approximately 20 GPU-hours, depending on the average generation length of the benchmark and model. 
Distilled reasoning models tend to require more time because they generate longer reasoning traces. 
The total compute scales approximately linearly with the number of evaluated model--benchmark settings and the number of sampled trajectories per question.

After trajectories and hidden states are cached, training the PAIR linear scorers is lightweight. 
The phase-specific logistic-regression models are trained on cached features and typically require only CPU-level computation or a small amount of CPU/GPU time compared with sampling. 
Steering also adds little overhead during generation, since it only applies a vector update to the hidden state at selected positions and does not require updating model parameters. 
The main resource bottlenecks are therefore generation time and storage for cached trajectories, token-level hidden states, and extracted step-level features.

\subsection{Supplementary MMLU Candidate-Set Analysis}
\label{app:mmlu_candidate_coverage}

This appendix-only analysis measures answer availability in a separate subject-balanced subset of 500 official MMLU test questions per model. It does not train or evaluate a PAIR scorer on MMLU. For each question, we retain 32 raw independent draws from Llama-3.1-8B-Instruct or Qwen3-4B, sampled with temperature 1.0 and top-$p$ 0.95 under a zero-shot multiple-choice chain-of-thought prompt; Qwen uses its native thinking mode. The first $N$ draws define each candidate set, with duplicate draws retained. Unparsed outputs count toward $N$ but cast no vote. Self-consistency (SC@$N$) chooses the most frequent parsed answer, breaking ties by the earliest valid draw; Oracle@$N$ is correct if any of the same $N$ candidates contains the correct answer. Thus Oracle is a candidate-set upper bound, not the accuracy of an implemented selector.

\begin{table}[htbp]
\centering
\caption{Self-consistency and oracle accuracy on the same fixed MMLU candidate sets. Each model uses 500 official-test questions; $N$ is the number of raw draws retained per question. Gap is Oracle minus SC in percentage points.}
\label{tab:mmlu_candidate_coverage}
\small\begin{tabular}{llrrr}
\toprule
Model & $N$ & SC@$N$ (\%) & Oracle@$N$ (\%) & Gap (pp) \\
\midrule
Llama-3.1-8B & 1 & 68.2 & 68.2 & 0.0 \\
Llama-3.1-8B & 4 & 72.4 & 85.0 & 12.6 \\
Llama-3.1-8B & 8 & 75.0 & 90.6 & 15.6 \\
Llama-3.1-8B & 16 & 76.4 & 95.4 & 19.0 \\
Llama-3.1-8B & 32 & 75.0 & 97.4 & 22.4 \\
\midrule
Qwen3-4B & 1 & 83.2 & 83.2 & 0.0 \\
Qwen3-4B & 4 & 84.4 & 90.8 & 6.4 \\
Qwen3-4B & 8 & 84.2 & 91.8 & 7.6 \\
Qwen3-4B & 16 & 84.6 & 93.0 & 8.4 \\
Qwen3-4B & 32 & 83.0 & 93.8 & 10.8 \\
\bottomrule
\end{tabular}
\end{table}

At $N=32$, the Oracle--SC gap is 22.4 percentage points for Llama and 10.8 points for Qwen; 95\% question-bootstrap intervals from 2,000 resamples are [19.0, 26.2] and [8.0, 13.8] points, respectively. These gaps quantify selection headroom within this MMLU candidate set; they do not imply that PAIR can close the gap. The multiple-choice prompt and dataset differ from the three main PAIR benchmarks, so these accuracies are not directly comparable to Table~\ref{tab:main_results}.
The SC curve need not increase monotonically with $N$: adding draws can change the majority-vote answer.

%% file: iclr2027_conference.bib
@misc{zhang2025reasoningmodelsknowtheyre,
      title={Reasoning Models Know When They're Right: Probing Hidden States for Self-Verification}, 
      author={Anqi Zhang and Yulin Chen and Jane Pan and Chen Zhao and Aurojit Panda and Jinyang Li and He He},
      year={2025},
      eprint={2504.05419},
      archivePrefix={arXiv},
      primaryClass={cs.AI},
      url={https://arxiv.org/abs/2504.05419}, 
}

@article{wei2022chain,
  title={Chain-of-thought prompting elicits reasoning in large language models},
  author={Wei, Jason and Wang, Xuezhi and Schuurmans, Dale and Bosma, Maarten and Xia, Fei and Chi, Ed and Le, Quoc V and Zhou, Denny and others},
  journal={Advances in neural information processing systems},
  volume={35},
  pages={24824--24837},
  year={2022}
}

@article{chen2021evaluating,
  title={Evaluating large language models trained on code},
  author={Chen, Mark and Tworek, Jerry and Jun, Heewoo and Yuan, Qiming and Pinto, Henrique Ponde De Oliveira and Kaplan, Jared and Edwards, Harri and Burda, Yuri and Joseph, Nicholas and Brockman, Greg and others},
  journal={arXiv preprint arXiv:2107.03374},
  year={2021}
}

@misc{he2026reasoningchainofthoughtlatentcomputational,
      title={Reasoning Beyond Chain-of-Thought: A Latent Computational Mode in Large Language Models}, 
      author={Zhenghao He and Guangzhi Xiong and Bohan Liu and Sanchit Sinha and Aidong Zhang},
      year={2026},
      eprint={2601.08058},
      archivePrefix={arXiv},
      primaryClass={cs.CL},
      url={https://arxiv.org/abs/2601.08058}, 
}

@article{sun2026llm,
  title={LLM Reasoning as Trajectories: Step-Specific Representation Geometry and Correctness Signals},
  author={Sun, Lihao and Dong, Hang and Qiao, Bo and Lin, Qingwei and Zhang, Dongmei and Rajmohan, Saravan},
  journal={arXiv preprint arXiv:2604.05655},
  year={2026}
}

@article{kojima2022large,
  title={Large language models are zero-shot reasoners},
  author={Kojima, Takeshi and Gu, Shixiang Shane and Reid, Machel and Matsuo, Yutaka and Iwasawa, Yusuke},
  journal={Advances in neural information processing systems},
  volume={35},
  pages={22199--22213},
  year={2022}
}

@article{wang2022self,
  title={Self-consistency improves chain of thought reasoning in language models},
  author={Wang, Xuezhi and Wei, Jason and Schuurmans, Dale and Le, Quoc and Chi, Ed and Narang, Sharan and Chowdhery, Aakanksha and Zhou, Denny},
  journal={arXiv preprint arXiv:2203.11171},
  year={2022}
}

@article{yao2023tree,
  title={Tree of thoughts: Deliberate problem solving with large language models},
  author={Yao, Shunyu and Yu, Dian and Zhao, Jeffrey and Shafran, Izhak and Griffiths, Tom and Cao, Yuan and Narasimhan, Karthik},
  journal={Advances in neural information processing systems},
  volume={36},
  pages={11809--11822},
  year={2023}
}

@inproceedings{besta2024graph,
  title={Graph of thoughts: Solving elaborate problems with large language models},
  author={Besta, Maciej and Blach, Nils and Kubicek, Ales and Gerstenberger, Robert and Podstawski, Michal and Gianinazzi, Lukas and Gajda, Joanna and Lehmann, Tomasz and Niewiadomski, Hubert and Nyczyk, Piotr and others},
  booktitle={Proceedings of the AAAI conference on artificial intelligence},
  volume={38},
  number={16},
  pages={17682--17690},
  year={2024}
}

@article{burns2212discovering,
  title={Discovering latent knowledge in language models without supervision, 2024},
  author={Burns, Collin and Ye, Haotian and Klein, Dan and Steinhardt, Jacob},
  journal={URL https://arxiv. org/abs/2212.03827}
}

@inproceedings{azaria2023internal,
  title={The internal state of an LLM knows when it’s lying},
  author={Azaria, Amos and Mitchell, Tom},
  booktitle={Findings of the Association for Computational Linguistics: EMNLP 2023},
  pages={967--976},
  year={2023}
}

@article{li2023inference,
  title={Inference-time intervention: Eliciting truthful answers from a language model},
  author={Li, Kenneth and Patel, Oam and Vi{\'e}gas, Fernanda and Pfister, Hanspeter and Wattenberg, Martin},
  journal={Advances in Neural Information Processing Systems},
  volume={36},
  pages={41451--41530},
  year={2023}
}

@article{park2023linear,
  title={The linear representation hypothesis and the geometry of large language models},
  author={Park, Kiho and Choe, Yo Joong and Veitch, Victor},
  journal={arXiv preprint arXiv:2311.03658},
  year={2023}
}

@article{liu2025llm,
  title={LLM Microscope: What Model Internals Reveal About Answer Correctness and Context Utilization},
  author={Liu, Jiarui and Jain, Jivitesh and Diab, Mona and Subramani, Nishant},
  journal={arXiv preprint arXiv:2510.04013},
  year={2025}
}

@article{turner2023activation,
  title={Activation addition: Steering language models without optimization. arXiv eprints, pages arXiv--2308},
  author={Turner, Alexander Matt and Thiergart, Lisa and Leech, Gavin and Udell, David and Vazquez, Juan J and Mini, Ulisse and MacDiarmid, Monte},
  year={2023}
}

@article{panickssery2023steering,
  title={Steering llama 2 via contrastive activation addition},
  author={Panickssery, Nina and Gabrieli, Nick and Schulz, Julian and Tong, Meg and Hubinger, Evan and Turner, Alexander Matt},
  journal={arXiv preprint arXiv:2312.06681},
  year={2023}
}

@article{boppana2026reasoning,
  title={Reasoning theater: Disentangling model beliefs from chain-of-thought},
  author={Boppana, Siddharth and Ma, Annabel and Loeffler, Max and Sarfati, Raphael and Bigelow, Eric and Geiger, Atticus and Lewis, Owen and Merullo, Jack},
  journal={arXiv preprint arXiv:2603.05488},
  year={2026}
}

@article{suzgun2022challenging,
  title={Challenging BIG-Bench Tasks and Whether Chain-of-Thought Can Solve Them},
  author={Suzgun, Mirac and Scales, Nathan and Sch{\"a}rli, Nathanael and Gehrmann, Sebastian and Tay, Yi and Chung, Hyung Won and Chowdhery, Aakanksha and Le, Quoc V and Chi, Ed H and Zhou, Denny and and Wei, Jason},
  journal={arXiv preprint arXiv:2210.09261},
  year={2022}
}

@article{cobbe2021gsm8k,
  title={Training Verifiers to Solve Math Word Problems},
  author={Cobbe, Karl and Kosaraju, Vineet and Bavarian, Mohammad and Chen, Mark and Jun, Heewoo and Kaiser, Lukasz and Plappert, Matthias and Tworek, Jerry and Hilton, Jacob and Nakano, Reiichiro and Hesse, Christopher and Schulman, John},
  journal={arXiv preprint arXiv:2110.14168},
  year={2021}
}

@article{lightman2023lets,
  title={Let's Verify Step by Step},
  author={Lightman, Hunter and Kosaraju, Vineet and Burda, Yura and Edwards, Harri and Baker, Bowen and Lee, Teddy and Leike, Jan and Schulman, John and Sutskever, Ilya and Cobbe, Karl},
  journal={arXiv preprint arXiv:2305.20050},
  year={2023}
}

@misc{meta_llama3_2024,
  author       = {Meta},
  title        = {Llama 3.1: Open Foundation and Instruction-Tuned Large Language Models},
  year         = {2024},
  url          = {https://ai.meta.com/llama/},
  note         = {Llama-3.1-8B-Instruct model}
}

@article{gemma_2025,
  title={Gemma 3 technical report},
  author={Team, Gemma and Kamath, Aishwarya and Ferret, Johan and Pathak, Shreya and Vieillard, Nino and Merhej, Ramona and Perrin, Sarah and Matejovicova, Tatiana and Ram{\'e}, Alexandre and Rivi{\`e}re, Morgane and others},
  journal={arXiv preprint arXiv:2503.19786},
  year={2025}
}

@misc{qwen3technicalreport,
      title={Qwen3 Technical Report}, 
      author={Qwen Team},
      year={2025},
      eprint={2505.09388},
      archivePrefix={arXiv},
      primaryClass={cs.CL},
      url={https://arxiv.org/abs/2505.09388}, 
}

@misc{deepseekai2025deepseekr1incentivizingreasoningcapability,
      title={DeepSeek-R1: Incentivizing Reasoning Capability in LLMs via Reinforcement Learning}, 
      author={DeepSeek-AI},
      year={2025},
      eprint={2501.12948},
      archivePrefix={arXiv},
      primaryClass={cs.CL},
      url={https://arxiv.org/abs/2501.12948}, 
}
